\documentclass[runningheads]{llncs}

\usepackage{eccv}

\usepackage{eccvabbrv}

\usepackage{graphicx}
\usepackage{booktabs}
\usepackage{ulem}
\usepackage{multirow}       
\usepackage[table]{xcolor}  

\usepackage[accsupp]{axessibility}  
\usepackage{hyperref}

\usepackage{orcidlink}
\newcommand{\Method}{S-VAM}

\begin{document}

\title{S-VAM: Shortcut Video-Action Model by Self-Distilling Geometric and Semantic Foresight}

\titlerunning{S-VAM}


\author{Haodong Yan\inst{1} \and Zhide Zhong\inst{1} \and Jiaguan Zhu\inst{1} \and Junjie He\inst{1} \and Weilin Yuan\inst{1} \and Wenxuan Song\inst{1} \and Xin Gong\inst{1} \and Yingjie CAI\inst{2} \and Guanyi Zhao\inst{2} \and Xu Yan\inst{2} \and Bingbing Liu\inst{2} \and Ying-Cong Chen\inst{1} \and Haoang Li\inst{1}}

\authorrunning{Yan, H. et al.}

\institute{The Hong Kong University of Science and Technology (Guangzhou) \and
Huawei Foundation Model Department}

\maketitle


\begin{abstract}
Video action models (VAMs) have emerged as a promising paradigm for robot learning, owing to their powerful visual foresight for complex manipulation tasks. However, current VAMs, typically relying on either slow multi-step video generation or noisy one-step feature extraction, cannot simultaneously guarantee real-time inference and high-fidelity foresight. To address this limitation, we propose S-VAM, a shortcut video-action model that foresees coherent geometric and semantic representations via a single forward pass. Serving as a stable blueprint, these foreseen representations significantly simplify the action prediction. To enable this efficient shortcut, we introduce a novel self-distillation strategy that condenses structured generative priors of multi-step denoising into
one-step inference. Specifically, vision foundation model (VFM) representations extracted from the diffusion model's own multi-step generated videos provide teacher targets. Lightweight decouplers, as students, learn to directly map noisy one-step features to these targets. Extensive experiments in simulation and the real world demonstrate that our S-VAM outperforms state-of-the-art methods, enabling efficient and precise manipulation in complex environments. Our project page is \url{https://haodong-yan.github.io/S-VAM/}
  \keywords{Robot Learning \and Video-Action Models \and Vision Foundation Models}
\end{abstract}


\section{Introduction}
Vision-Language-Action (VLA) models~\cite{zitkovich2023rt, black2024pi_0,intelligence2025pi_,kim2025openvla,song2025reconvla, song2025pd} have emerged as a dominant paradigm for robotic manipulation. By equipping pretrained vision-language models (VLMs) with action heads and finetuning on robot action data, these models learn to predict executable actions from visual observations and instructions. However, standard VLMs are pre-trained on static image-text pairs, inherently lacking the spatiotemporal foresight essential for dynamic interaction. Consequently, VLA models must learn physical dynamics directly from robot action data. This reliance causes a critical bottleneck, as high-quality robot action data is significantly more expensive to collect compared to abundant web-scale video data, thereby severely limiting the scalability of direct VLA approaches~\cite{li2025mimicdreamer}.

To address the above challenges, recent work~\cite{huvideo,xie2025human2robot,shen2025videovla,chen2025large,lingbot-va2026,black2024zeroshot,du2023learning,feng2025vidar} has increasingly adopted a video-action model (VAM) pipeline: a video diffusion model (VDM) synthesizes a visual plan conditioned on the current observation, in tandem with an action expert (AE) predicting the corresponding control signals. This design leverages the predictive priors learned from abundant internet-scale videos within the VDM, which enables the AE to learn precise policies while relying on substantially less robot action data.

\begin{figure}[!t]
    \centering
    \includegraphics[width=.98\linewidth]{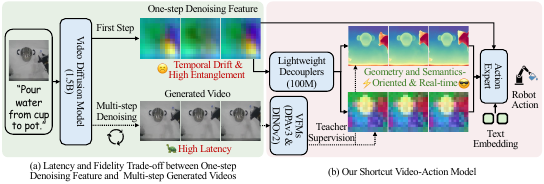}
    \caption{Motivation and Overview of our Shortcut Video-Action Model.
(a) Current video-action models struggle with a trade-off: \textit{one-step feature extraction} is fast but yields noisy and entangled representations, whereas \textit{multi-step video generation} predicts precise future states but is too slow for real-time control.
(b) To address this, we propose a shortcut video-action model that foresees coherent geometric and semantic representations via a single forward pass. Specifically, we introduce a self-distillation strategy that extracts vision foundation model (VFM) representations (DPAv3~\cite{lin2025depth} and DINOv2~\cite{oquab2023dinov2}) from the diffusion model's own multi-step generated videos to serve as teacher supervision exclusively during training (dashed path). By employing lightweight decouplers as students to map entangled one-step features to these geometry and semantics-oriented targets, our approach condenses structured generative priors of multi-step denoising into one-step inference, thereby enabling real-time and precise action prediction.}
\label{fig:teaser}

\end{figure}

As illustrated in Fig.~\ref{fig:teaser}(a), current VAMs mainly fall into two paradigms.
The first paradigm~\cite{black2024zeroshot,du2023learning,feng2025vidar, dreamzero2025, kim2026cosmos} relies on generating future videos to guide control. While these videos provide detailed planning priors, the prohibitive latency of multi-step denoising renders it unsuitable for high-frequency closed-loop control. To enable efficient inference, the second paradigm (e.g., Video Prediction Policy (VPP)~\cite{huvideo}) utilizes one-step denoising features, driven by the insight that initial denoising steps focus on establishing a global future blueprint, while later steps refine local details. However, these one-step features remain noisy and highly entangled, exhibiting poor temporal coherence (e.g., the representation of gripper drifts across frames shown in Fig.~\ref{fig:teaser}(a)).
Such entangled representations lack geometry-oriented cues needed to compensate for monocular depth ambiguity and the semantic distinctiveness required to distinguish task-relevant objects from irrelevant elements in the scene.

To overcome the above dilemma between inference latency and foresight fidelity in existing VAMs, we propose S-VAM, a shortcut video-action model that foresees coherent geometric and semantic representations via a single forward pass for guiding precise action generation. As shown in Fig.~\ref{fig:teaser}(b), our S-VAM employs relatively lightweight decouplers to explicitly disentangle noisy one-step diffusion features into specialized vision foundation model (VFM) representations. Crucially, to enable this shortcut, we introduce a novel self-distillation strategy. During training, the decouplers serve as student networks, directly supervised by stable VFM representation targets extracted from videos generated by the VDM via multi-step denoising. Specifically, we select DPAv3~\cite{lin2025depth} to provide dynamic geometry-oriented targets, and DINOv2~\cite{oquab2023dinov2} to yield patch-level semantic targets. This self-distillation process condenses structured generative priors of multi-step denoising into one-step inference to drive efficient and precise robotic manipulation.
To summarize, our contributions are as follows:

\begin{itemize}
  \item We propose S-VAM, a shortcut video-action model that foresees coherent geometric and semantic representations in a single feed-forward pass to drive precise action generation.
  \item To enable this efficient shortcut, we introduce a self-distillation strategy where lightweight decouplers serve as students to directly map entangled one-step denoising features to geometry and semantics-oriented VFM teacher targets derived from multi-step generated videos.
  \item Extensive experiments in simulation and the real world demonstrate that our S-VAM outperforms state-of-the-art methods, enabling efficient and precise manipulation in complex scenarios.
\end{itemize}

\section{Related Works}
\subsection{Vision-Language-Action Models}

Vision-Language-Action models~\cite{brohan2022rt,zitkovich2023rt,song2025reconvla,lin2025hif,li2024visionlanguage,kim2025openvla} establish an end-to-end mapping from text instructions and visual observations to executable actions. A prevailing paradigm in recent VLA research involves fine-tuning pre-trained VLMs on robot action data. Representative frameworks such as RT-2~\cite{zitkovich2023rt}, OpenVLA~\cite{kim2025openvla}, $\pi_{0}$~\cite{black2024pi_0}, $\pi_{0.5}$~\cite{intelligence2025pi_}, and CogVLA~\cite{li2025cogvla} leverage this strategy to achieve promising performance across diverse manipulation tasks. However, these methods inherently struggle to capture the spatiotemporal foresight essential for dynamic interaction, as their underlying VLMs are primarily pretrained on static image-text pairs.
Distinct from standard VLA paradigms, we augment the action model with pre-trained video generation models that inherently encode rich spatiotemporal dynamics and physical laws from internet-scale video datasets.

\subsection{Video Generation Models for Robot Learning}
Recent research~\cite{finn2016unsupervised,zhou2024robodreamer,huvideo,xie2025human2robot,kim2026cosmos,pai2025mimicvideovideoactionmodelsgeneralizable,black2024zeroshot,du2023learning,feng2025vidar} has increasingly explored leveraging video generation models for their capability of visual foresight. A common approach~\cite{NEURIPS2023_46a12649, black2024zeroshot,du2023learning,feng2025vidar} guides action generation by synthesizing future videos. Under this framework, a VDM drafts a visual plan from the observation alongside an action expert that outputs control signals. Despite their effectiveness, these methods suffer from prohibitive inference latency due to the multi-step denoising iterations required for high-quality generation. To address this issue, some works~\cite{huvideo,xie2025human2robot} directly extract one-step denoising features from video generation models to serve as future representations. However, these one-step features are noisy and highly entangled, which hinders precise action prediction. To unlock the full potential of these informative but entangled features, we introduce a shortcut that maps the one-step features to coherent geometric and semantic representations, achieving efficient inference without compromising the high-fidelity planning priors essential for precise control.

\subsection{Vision Foundation Models in Robot Learning}

VFMs have been extensively adopted in the field of robot learning. A prevalent paradigm~\cite{kim2025openvla,song2025reconvla,lin2025hif,black2024pi_0} leverages DINO~\cite{oquab2023dinov2,simeoni2025dinov3} and SigLIP~\cite{zhai2023sigmoid} encoders to capture fine-grained local and global semantic context from observation frames. To further enhance geometric understanding, several approaches incorporate explicit 3D cues. For instance, SpatialVLA~\cite{qu2025spatialvla} utilizes off-the-shelf depth estimators to generate pseudo point clouds as input, and DepthVLA~\cite{yuan2025depthvla} leverages estimated depth maps to train a dedicated depth expert for geometric reasoning. Distinct from these explicit injection methods, Spatial Forcing~\cite{li2025spatial} proposes a simple yet effective alignment strategy that implicitly compels VLA models to acquire geometry-aware knowledge. However, these methods are typically restricted to encoding or aligning the \textit{ current} static observation, missing the critical capability of \textit{future} visual foresight. In contrast, our shortcut video-action model directly foresees future geometric and semantic VFM representations.

\section{Method}
\label{sec:method}

\begin{figure}[t]
  \centering
  \includegraphics[width=0.99\linewidth]{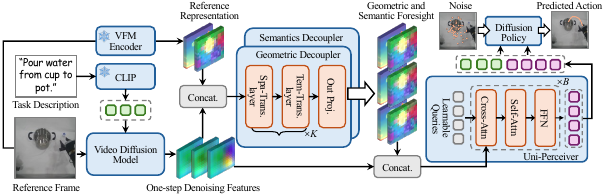}
 \caption{Architecture of our S-VAM. The core technical novelty lies in establishing a \textbf{shortcut} that bypasses the prohibitive latency of iterative video generation. Specifically, specialized decouplers disentangle highly entangled one-step diffusion features into coherent geometric and semantic foresight. This foresight is then aggregated with original features by a Uni-Perceiver, providing a holistic condition context for the downstream diffusion policy to predict precise robot action.} 
  \label{fig:method_overview}
\end{figure}

As shown in \cref{fig:method_overview}, we propose S-VAM, a shortcut video-action model that foresees geometric and semantic representations for action generation without multi-step denoising. Specifically, we first introduce the preliminaries in \cref{sec:preliminaries}. We then detail geometric and semantic decouplers in \cref{sec:distill}, which disentangle noisy one-step denoising features into future VFM representations. Finally, in \cref{sec:ae}, we describe how the Uni-Perceiver module fuses these representations with original diffusion features to condition the downstream diffusion policy.

\subsection{Preliminaries}
\label{sec:preliminaries}

\noindent \textbf{Video Diffusion Models.}
Our framework leverages Stable Video Diffusion (SVD)~\cite{blattmann2023stable} for the video generation backbone. SVD utilizes a VAE to encode the input video $V \in \mathbb{R}^{T \times 3 \times H \times W}$ into a compressed latent representation $z \in \mathbb{R}^{T \times C \times h \times w}$. Operating within this latent space, the model generates a video by iteratively removing Gaussian noise from a completely noisy latent $z_S \sim \mathcal{N}(0, \mathbf{I})$ through a reverse denoising process:
\begin{equation}
    z_{s-1} = \frac{1}{\sqrt{\alpha_s}} \left( z_s - \frac{1-\alpha_s}{\sqrt{1-\bar{\alpha}_s}} \epsilon_\theta(z_s, s, P,I) \right) + \sigma_s\epsilon,
\end{equation}
where $\epsilon_\theta(\cdot)$ is a learnable noise estimator and $\epsilon_\theta(z_s, s, P, I)$ is the predicted noise conditioned on the observed frame $I$ and task description $P$. However, such multi-step sampling leads to high inference latency that hinders real-time control.

\noindent \textbf{Vision Foundation Models.}
Pre-trained on internet-scale datasets, VFMs have demonstrated remarkable capabilities in learning highly structured and regularized representation spaces. We leverage frozen VFMs to extract these representations from the model's own multi-step generated video frames $\hat{V}$ to serve as teacher targets for self-distilling geometric and semantic foresight (see~\cref{sec:distill}). 
Formally, given a VFM, we extract the target representation $Y \in \mathbb{R}^{T\times C_\textnormal{VFM} \times h \times w}$ as:
\begin{equation}
    Y = \Phi(\operatorname{Interpolate}(\hat{V})),
\label{eq:vfm}
\end{equation}
where $\Phi(\cdot)$ denotes the VFM encoder and $\operatorname{Interpolate}(\cdot)$ is the linear interpolation operation that aligns the output feature maps to the target spatial resolution $h \times w$. Here, $C_\textnormal{VFM}$ denotes the output channel dimension determined by the specific VFM. 

\subsection{Geometric and Semantic Foresight Distillation}
\label{sec:distill}
\noindent \textbf{One-step Denoising Features.}
To enable efficient inference, we extract predictive features via a single denoising forward pass following~\cite{huvideo}. Specifically, we leverage multi-layer intermediate features from the up-sampling blocks. While inherently noisy and entangled, these internal representations are highly informative, thereby providing essential cues to be distilled into coherent geometric and semantic foresight.

Let $F_l \in \mathbb{R}^{T \times C_l \times h_l \times w_l}$ denote the feature map from the $l^{th}$ up-sampling layer of denoising network $\epsilon_\theta(Z_{S},S,P,I)$ at the first denoising step $S$. To effectively aggregate these features across different resolutions, we align each layer to a unified target spatial dimension $h \times w$ via linear interpolation $\operatorname{Interpolate}(\cdot)$:
\begin{equation}
    F'_l = \operatorname{Interpolate}(F_l), \quad F'_l \in \mathbb{R}^{T \times C_l \times h \times w}.
\end{equation}
The final comprehensive one-step feature $F$ is constructed by concatenating these aligned layers along the channel dimension:
\begin{equation}
    F = \operatorname{Concat}((F'_0, \dots, F'_L), \text{dim}=1), \quad F \in \mathbb{R}^{T \times C_{\Sigma} \times h \times w},
\end{equation}
where $C_{\Sigma} = \sum_{l=0}^{L} C_{l}$ denotes the total channel dimension of the aggregated features.

\noindent \textbf{Geometric and Semantic Decouplers.}
To exploit the rich but entangled dynamics within one-step features $F$, we introduce the geometric and the semantic decouplers that directly disentangle these noisy one-step features into coherent geometric and semantic VFM representations. Specifically, the geometric branch targets the representation space of DPAv3~\cite{lin2025depth}, and the semantic branch targets that of DINOv2~\cite{oquab2023dinov2}. This design choice is motivated by their complementary strengths: DPAv3 provides dynamic geometric structures, and DINOv2 offers patch-level semantic distinctiveness.

In terms of architecture, we instantiate each decoupler as a spatio-temporal transformer. For each branch $i \in \{\textnormal{geo}, \textnormal{sem}\}$, instead of inferring future states solely from the noisy diffusion features, the module leverages the reference representation of the current observation, $Y_{i}^{\textnormal{ref}} = \Phi_{i}(\textnormal{Interpolate}(I_\textnormal{obs}))\in\mathbb{R}^{1\times{C_i}\times h \times w}$, as a strong conditioning anchor. This anchor eases the optimization burden, effectively guiding the network to map unstable diffusion features into coherent VFM representations.
Formally, we temporally replicate the reference representation $Y_{\textnormal{ref}}^{i}$ to match the sequence length of the one-step feature volume $F$, and concatenate them along the channel dimension to obtain the fused input representation:
\begin{equation}
    \tilde{F}_i^0 = \text{Concat}((F, \text{Repeat}(Y_{i}^{\textnormal{ref}})), \text{dim}=1), \quad i \in \{\textnormal{geo}, \textnormal{sem}\}. 
\end{equation}
Then, this fused representation $\tilde{F}_i^0$ is first projected into a compact latent space via a linear embedding layer ($\mathbb{R}^{T\times (C_{i}+C_{\Sigma}) \times h \times w} \rightarrow \mathbb{R}^{T\times C_\textnormal{hidden} \times h \times w}$). The compressed features are then contextualized through a stack of $K$ factorized spatio-temporal blocks. Specifically, for each block $k$ ($1 \le k \le K$), the features are processed sequentially by spatial and temporal aggregation:
\begin{equation}
    \tilde{F}_i^k = \mathcal{T}_i^k \left( \mathcal{S}_i^k ( \tilde{F}_i^{k-1} ) \right), \quad i \in \{\textnormal{geo}, \textnormal{sem}\},
\end{equation}
where $\mathcal{S}(\cdot)$ and $\mathcal{T}(\cdot)$ denote the spatial and temporal transformer layers, respectively. Finally, the output from the $K^{th}$ block, $\tilde{F}_i^K$, is mapped back to the target VFM dimension via an output projection ($\mathbb{R}^{T\times C_\textnormal{hidden} \times h \times w} \rightarrow \mathbb{R}^{T\times C_i \times h \times w}$).

\noindent \textbf{Self-Distillation Optimization Objectives.}
As indicated by the dashed lines in \cref{fig:teaser}(b), we apply specific supervision to each branch by introducing a self-distillation mechanism, where VFM representations extracted from the diffusion model's own multi-step generated video $\hat{V}$ serve as teacher signals. The core advantage of this self-distillation is that the generated video originates from the same diffusion trajectory as the one-step features. In contrast, relying on targets extracted from ground-truth future frames would introduce an inherent trajectory misalignment that hinders optimization. Given that geometric and semantic branches target distinct representation spaces, we optimize them independently. Utilizing the extraction process in~\cref{eq:vfm} to obtain the teacher representations $Y_{i}$, we formulate the optimization objective for each branch $i \in \{\textnormal{geo}, \textnormal{sem}\}$:
\begin{equation}
    \mathcal{L}_{i} = \| \tilde{F}_{i}^K - Y_{i} \|_2^2, \quad i \in \{\textnormal{geo}, \textnormal{sem}\}. 
    \label{eq:distill_loss}
\end{equation}
By minimizing $\mathcal{L}_\textnormal{geo}$ and $\mathcal{L}_\textnormal{sem}$, our S-VAM effectively condenses the multi-step generation capability into our single-step shortcut, yielding high-fidelity foresight without the latency of iterative denoising.

\subsection{Action Expert}
\label{sec:ae}
To bridge the gap between decoupled geometric and semantic foresight and physical control, our action expert employs a Uni-Perceiver to aggregate a holistic context, guiding a diffusion policy to predict actions.

\noindent \textbf{Uni-Perceiver.}
We first construct a holistic conditioning context, denoted as $\mathcal{C}=\textnormal{Concat}(\tilde{F}^{K}_\textnormal{geo}, \tilde{F}^{K}_\textnormal{sem}, F)\in\mathbb{R}^{T\times C_{\textnormal{hol}} \times h \times w}$, by concatenating the geometric and semantic foresight with the original one-step diffusion feature along the feature channel dimension. Retaining the original feature $F$ is critical, as it preserves residual global context that complements the explicit geometric and semantic foresight.
To mitigate the curse of dimensionality and the associated computational overhead introduced by $\mathcal{C}$, we compress it into a highly condensed token representation $F_\textnormal{agg} \in \mathbb{R}^{N \times C_\textnormal{agg}}$. This condensation effectively filters out redundant information to facilitate the diffusion policy in learning to predict precise actions. Specifically, we initialize a compact set of $N$ learnable latent queries $\mathcal{Q}$ that aggregate spatiotemporal information through a QFormer-style~\cite{li2023blip,jaegle2022perceiver} token condensation module:
\begin{equation}
    F_\textnormal{agg} = \text{FFN}(\text{SelfAttn}(\text{CrossAttn}(\mathcal{Q}, \mathcal{C}))).
    \label{eq:aggregation}
\end{equation}
Here, the cross-attention layer $\textnormal{CrossAttn}(\cdot)$ aggregates salient spatiotemporal features from the holistic context $\mathcal{C}$ into latent queries $\mathcal{Q}$, while the subsequent self-attention layer $\textnormal{SelfAttn}(\cdot)$ models the internal interactions among these $N$ condensed tokens. Finally, a feed-forward network $\textnormal{FFN}(\cdot)$ further enhances the expressiveness of the condensed tokens.

\noindent \textbf{Diffusion Policy.}
Finally, we incorporate the aggregated representation $F_\textnormal{agg}$ and the text embedding $E$ of the task description into a Diffusion Transformer (DiT) architecture via cross-attention layers. The policy is trained to reconstruct the ground-truth action sequence $a_0$ from the pure noise $a_J$ by predicting the added noise $\epsilon$. The training objective is formulated as follows:
\begin{equation}
    \mathcal{L}_A = \mathbb{E}_{j, a_j, \epsilon} \left[ \left\| \epsilon - \epsilon_\phi(a_j, F_\textnormal{agg}, E, j) \right\|_2^2 \right],
\end{equation}
where $a_j$ is the noisy action at diffusion timestep $j$, and $\epsilon_\phi(\cdot)$ denotes the noise prediction network.

\section{Experiment}

We evaluate our \Method~on both simulated and real-world robotic tasks to answer the following questions: (\textbf{Q1}) Can \Method~outperform state-of-the-art methods? (\textbf{Q2}) How does each key component of \Method~(e.g., geometric distillation, semantic distillation, and Uni-Perceiver) contribute to the overall performance? (\textbf{Q3}) How do alternative VFM representations affect distillation and downstream control? (\textbf{Q4}) Can \Method~adapt to complex real-world scenarios? 

\subsection{Experimental Setup}
\noindent \textbf{Simulated Benchmarks.} 
We evaluate our S-VAM on two challenging manipulation benchmarks. 1) CALVIN~\cite{mees2022calvin} ABC$\rightarrow$D includes four distinct tabletop environments (ABCD). Policies are trained on ABC and evaluated on unseen D to assess generalization across consecutive long-horizon tasks. 2) MetaWorld~\cite{yu2020meta} features a Sawyer robot performing 50 manipulation tasks across varying difficulty levels as defined in~\cite{seo2023masked}. The training set consists of 50 demonstrations for each task. For fair comparison, all experiments are conducted following~\cite{huvideo} under a third-view setup using only the primary monocular RGB camera.

\noindent \textbf{Implementation Details.}
Our pipeline consists of a three-stage training process.
In the first stage, we fine-tune the SVD backbone, initialized from weights pretrained on embodied scenes~\cite{huvideo}.
We train for 100k steps on MetaWorld and 40k steps on real-world tasks using 4 NVIDIA H100 GPUs.
For CALVIN, we directly adopt the fine-tuned model from~\cite{huvideo} without additional training.
In the second stage, we freeze the video generation model and train the geometric and semantic decouplers to distill coherent foresight from noisy one-step features. This self-distillation is efficient, requiring only 50k steps on a single NVIDIA H100 GPU for all benchmarks.
In the final stage, while keeping the SVD backbone and decouplers frozen, we exclusively train the action expert to decode control signals. We train this stage for 60k steps on CALVIN and 40k steps on the remaining benchmarks, using four NVIDIA H100 GPUs. For inference, we use a single NVIDIA RTX 3090 (24GB) GPU. In qualitative analysis, we additionally train a DPT head~\cite{Ranftl_2021_ICCV} that maps geometric foresight to depth maps following a standard probing protocol~\cite{li2025spatial,wu2026geometry}. 

\noindent \textbf{Methods for Comparison.} 
We compare our S-VAM against state-of-the-art baselines, broadly categorized into two paradigms. \textbf{Direct action learning} methods (e.g., RT-1~\cite{brohan2022rt}, Diffusion Policy~\cite{chi2025diffusion}, OpenVLA~\cite{kim2025openvla}, CLOVER~\cite{bu2024closedloop}, $\pi_0$~\cite{black2024pi_0}, and Spatial Forcing~\cite{li2025spatial}) directly map observations to actions without explicitly modeling future dynamics, whereas \textbf{predictive methods} (e.g., SuSIE~\cite{black2024zeroshot}, GR-1~\cite{wu2023unleashing}, VPP~\cite{huvideo}, Uni-VLA~\cite{wang2025unified}, and HiF-VLA~\cite{lin2025hif}) predict future states or intermediate representations to guide action generation.
\subsection{Evaluations Results on Simulated Benchmarks (Q1)}

\begin{figure}[!t]
  \centering
  \includegraphics[width=1\linewidth]{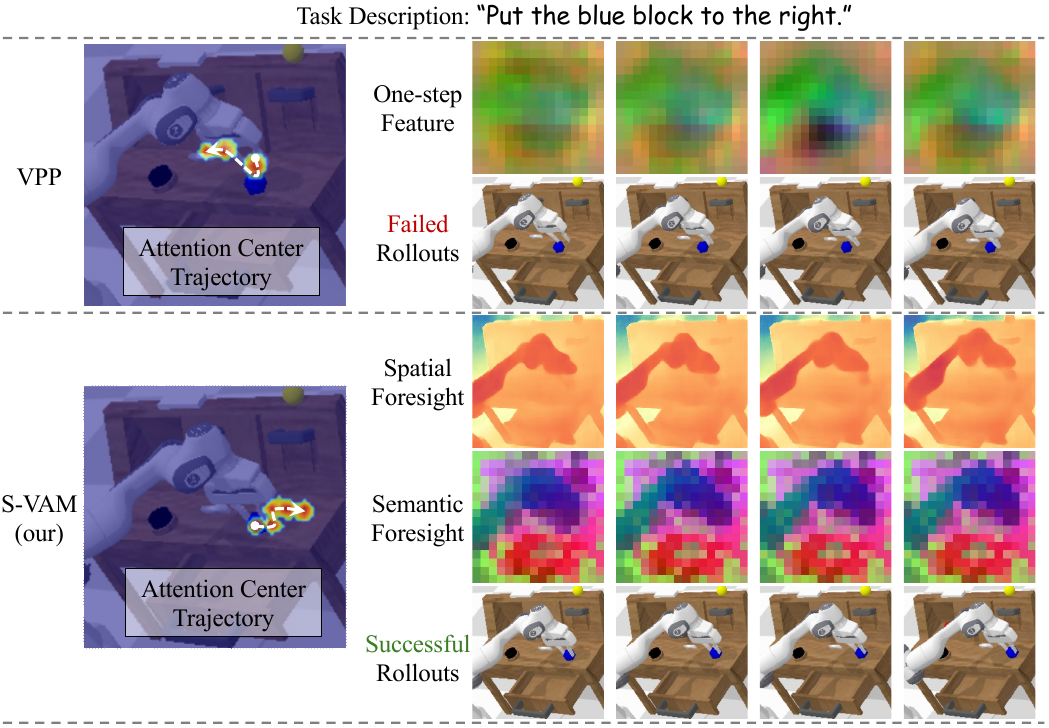}

\caption{Qualitative comparison on CALVIN~\cite{mees2022calvin}. VPP~\cite{huvideo} utilizes entangled one-step features, resulting in an erratic attention trajectory that explicitly contradicts the language instruction and leads to failed actuation. In contrast, our S-VAM foresees geometric and semantic representations. This decoupled future blueprint enables the action expert to anchor a coherent attention trajectory that perfectly aligns with the language instruction, thereby ensuring successful execution. \textit{Note: Geometric foresight is visualized as probe-based depth maps}~\cite{li2025spatial,wu2026geometry}\textit{; one-step and semantic features are visualized via PCA computed globally across the entire sequence.}}
\label{fig:qualitative_viz}

\end{figure}

\begin{table}[!t]
\centering
\small
\caption{Quantitative comparison on CALVIN~\cite{mees2022calvin}. We report stage-wise success rates (1-5) and average sequence length (Avg. Len.). \textbf{Best} and \underline{second-best} results are highlighted accordingly.}
\label{tab:results}
\renewcommand{\arraystretch}{1.00}
\definecolor{mygray}{gray}{0.95}  
\scalebox{0.95}{
\begin{tabular}{c|c|ccccc|c}
\toprule
\multirow{2}{*}{\textbf{Category}} & \multirow{2}{*}{\textbf{Method}} & \multicolumn{5}{c|}{\textbf{$i^{th}$ Task Success Rate}} & \multirow{2}{*}{\textbf{Avg. Len.} $\uparrow$} \\
\cmidrule(lr){3-7} 
 &  & \textbf{1} & \textbf{2} & \textbf{3} & \textbf{4} & \textbf{5} & \\
\midrule

\multirow{4}{*}{\shortstack{Direct Action \\
Learning Methods}} 
 & OpenVLA~\cite{kim2025openvla}   & 91.3 & 77.8 & 62.0 & 52.1 & 43.5 & 3.27 \\
 & CLOVER~\cite{bu2024closedloop}    & \textbf{96.0} & 83.5 & 70.8 & 57.5 & 45.4 & 3.53 \\
 & $\pi_0$~\cite{black2024pi_0}   & 93.7 & 83.2 & 74.0 & 62.9 & 51.0 & 3.65 \\
 & Spatial Forcing~\cite{li2025spatial}  &93.6 &85.8 &78.4 &72.0& 64.6& 3.94 \\

\midrule
 & SuSIE~\cite{black2024zeroshot}  & 87.0 & 69.0 & 49.0 & 38.0 & 26.0 & 2.69 \\
 & VPP~\cite{huvideo}    & 90.9 & 81.5 & 71.3 & 62.0 & 51.8 & 3.58 \\

 &Uni-VLA~\cite{wang2025unified} & 95.5& 85.8& 74.8& 66.9& 56.5& 3.80 \\
  & HiF-VLA~\cite{lin2025hif}  & 93.5 &\underline{87.4} & \underline{81.4 }& \underline{75.9} & \textbf{69.4} & \underline{4.08} \\

\rowcolor{mygray}
\cellcolor{white} \multirow{-5}{*}{\shortstack{Predictive \\  Methods}} 
&\textbf{S-VAM(our)}  & \underline{95.8} & \textbf{90.7} & \textbf{83.7} & \textbf{77.0} & \underline{68.9} & \textbf{4.16}\\

\bottomrule
\end{tabular}
}
\end{table}

\noindent \textbf{Results on CALVIN~\cite{mees2022calvin}.} 
As demonstrated in \cref{tab:results}, our S-VAM achieves a state-of-the-art average length of \textbf{4.16}, outperforming both direct action learning and predictive baselines. Notably, our S-VAM surpasses our direct baseline, VPP~\cite{huvideo}, by a significant margin of 0.58. As shown in \cref{fig:qualitative_viz}, VPP~\cite{huvideo} relies on entangled one-step features, which leads to an erratic attention center trajectory\footnote{Using head-averaged attention maps from the final cross-attention layer of the token condensation module, we extract this trajectory by tracking the maximum activation (and its 3-pixel neighborhood) across frames.} that explicitly contradicts the language instruction, resulting in imprecise actuation. In contrast, by distilling decoupled geometric and semantic foresight, our S-VAM anchors a coherent attention trajectory that perfectly aligns with the language instruction. Our execution rollouts strictly follow this high-fidelity blueprint, confirming that the action expert successfully translates these future representations into precise manipulation. Furthermore, our S-VAM outperforms Spatial Forcing~\cite{li2025spatial} by distilling future states instead of merely aligning representations on observation frames, enabling the agent to better anticipate complex interactions. Additionally, our S-VAM surpasses other predictive HiF-VLA~\cite{lin2025hif} and Uni-VLA~\cite{wang2025unified}. Although they also employ predictive mechanisms, their underlying VLMs inherently lack the rich spatiotemporal dynamics naturally captured by our video backbone.

\begin{figure}[!t]
  \centering
  \includegraphics[width=1\linewidth]{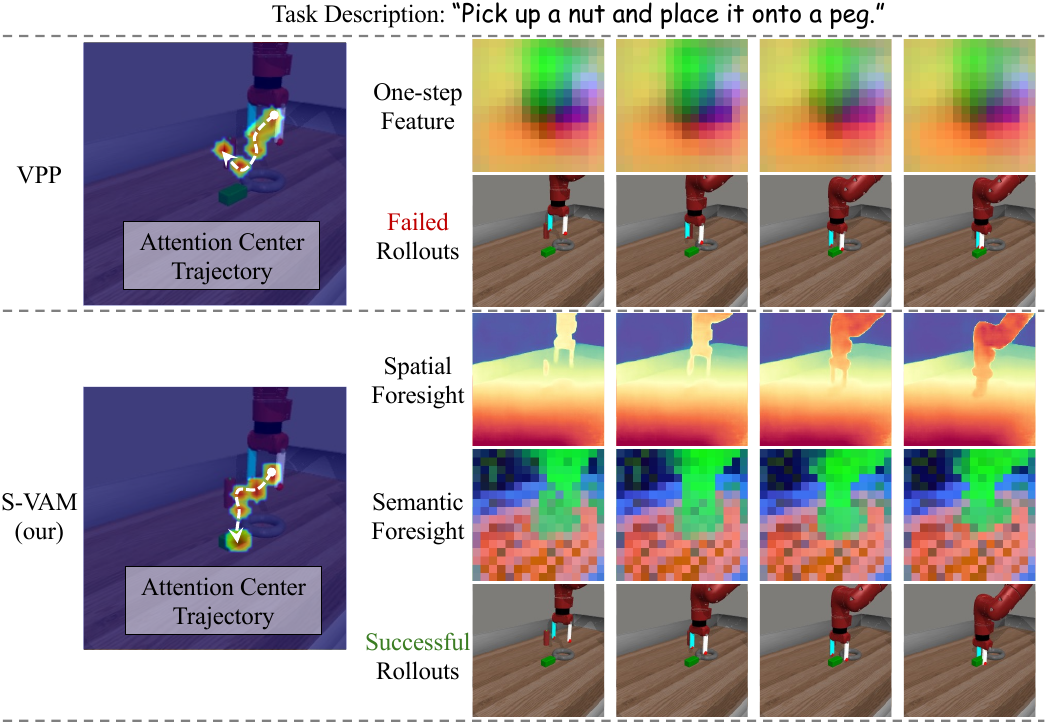}
\caption{
Qualitative comparison on MetaWorld~\cite{yu2020meta}.
VPP~\cite{huvideo} utilizes entangled one-step features, resulting in a diverging attention trajectory that completely misses the target ``nut''. In contrast, our S-VAM foresees explicit geometric and semantic representations. This decoupled future blueprint enables the action expert to anchor a coherent attention trajectory that accurately guides to the instructed target, ensuring successful grasping. \textit{Note: Visualization settings follow~\cref{fig:qualitative_viz}.}
}
\label{fig:qualitative_viz_meta}
\end{figure}
\begin{table}[!t]
\centering
\small
\caption{Quantitative comparison on Metaworld~\cite{yu2020meta}. We report success rates on easy, middle, and hard tasks, along with the overall average. \textbf{Best} and \underline{second-best} results are highlighted accordingly.}
\label{tab:metaworld_results}

\definecolor{mygray}{gray}{0.95}

\renewcommand{\arraystretch}{1.00}
\scalebox{0.95}{
\begin{tabular}{c|cccc|c}

\toprule
\multirow{2}{*}{\textbf{Category}} & \multirow{2}{*}{\textbf{Method}} & \textbf{Easy} & \textbf{Middle} & \textbf{Hard} & \textbf{Average} $\uparrow$ \\
 & & (28 tasks) & (11 tasks) & (11 tasks) & (50 tasks) \\
\midrule

\multirow{3}{*}{\shortstack{Direct Action \\
Learning Methods}} 
 & RT-1~\cite{brohan2022rt}           & 0.605  & 0.042 & 0.015  & 0.346 \\
 & Diffusion Policy~\cite{chi2025diffusion} & 0.442 & 0.062 & 0.095 & 0.279 \\
 & Spatial Forcing~\cite{li2025spatial} & 0.737 & 0.436 & 0.451 & 0.609 \\
 
\midrule

 & SuSIE~\cite{black2024zeroshot}  & 0.560 & 0.196 & 0.255 & 0.410 \\
 & GR-1~\cite{wu2023unleashing}  & 0.725 & 0.327 & 0.451 & 0.574 \\
 & HiF-VLA~\cite{lin2025hif} & 0.729 & 0.364 & 0.404 & 0.577 \\
 & VPP~\cite{huvideo} & \textbf{0.818} & \underline{0.493} & \underline{0.526} & \underline{0.682} \\
 

\rowcolor{mygray}
\cellcolor{white} \multirow{-5}{*}{\shortstack{Predictive \\  Methods}} 
& \textbf{S-VAM (our)} & \underline{0.793} & \textbf{0.607} & \textbf{0.684} & \textbf{0.728} \\

\bottomrule
\end{tabular}
}
\end{table}
\noindent \textbf{Results on MetaWorld~\cite{yu2020meta}.} 
As shown in~\cref{tab:metaworld_results}, our S-VAM achieves the highest average success rate of \textbf{72.8\%}, surpassing the previous state-of-the-art (SOTA) method VPP~\cite{huvideo} by a clear margin. The advantage of our S-VAM is particularly demonstrated in hard tasks, where it achieves a success rate of 68.4\%, significantly outperforming VPP ($52.6\%$). Hard tasks in MetaWorld often involve complex object interactions and fine-grained geometric constraints, such as the precise assembly visualized in \cref{fig:qualitative_viz_meta}. In this task, the failure of VPP~\cite{huvideo} largely stems from its entangled one-step features, which produce a diverging attention trajectory that completely misses the target ``nut''. In contrast, the decoupled geometric and semantic foresight enables our S-VAM to anchor a coherent attention trajectory that accurately traces a path to the instructed target for a precise grasp. Furthermore, S-VAM surpasses leading non-VAM methods under the 50-demonstration setting, outperforming Spatial Forcing~\cite{li2025spatial} and HiF-VLA~\cite{lin2025hif} by 11.9\% and 15.1\%, respectively. This underscores the profound advantage of leveraging dynamic priors from the video generation model in data-scarce scenarios.

\begin{table}[!t]
\centering
\small
\caption{Ablation study on key components of our S-VAM.}
\label{tab:ablation}

\definecolor{mygray}{gray}{0.95}

\renewcommand{\arraystretch}{1.00}
\scalebox{0.95}{
\begin{tabular}{l|ccccc|c}
\toprule
\multirow{2}{*}{\textbf{Method Variant}} & \multicolumn{5}{c|}{\textbf{$i^{th}$ Task Success Rate}} & \multirow{2}{*}{\textbf{Avg. Len.} $\uparrow$} \\
\cmidrule(lr){2-6}
 & \textbf{1} & \textbf{2} & \textbf{3} & \textbf{4} & \textbf{5} & \\
\midrule

w/o Geometric Distillation & 94.1 & 87.1 & 79.3 & 73.5 & 66.5 & 4.01 \\

w/o Semantic Distillation & 94.0 & 87.1 & 80.4 & 73.2 & 64.1 & 3.99 \\

w/o Self-Distillation   & 94.2 & 85.2 & 75.9 & 67.8 & 59.0 &  3.82\\
\midrule
w/o Uni-Perceiver & 94.0 & 84.6 & 74.7 & 65.1 & 53.8 & 3.72\\

w/o Original Diffusion Feature & 95.3 & 86.6 & 77.6 & 70.9 & 62.5 & 3.93\\


\midrule
\rowcolor{mygray}
\textbf{S-VAM (Full)} & \textbf{95.8} & \textbf{90.7} & \textbf{83.7} & \textbf{77.0} & \textbf{68.9} & \textbf{4.16} \\

\bottomrule
\end{tabular}
}
\end{table}

\subsection{Ablation Study (Q2)}
\label{sec:ablation}

To validate the effectiveness of key components in our S-VAM, we conduct an ablation study on the CALVIN~\cite{mees2022calvin} benchmark. Results are summarized in \cref{tab:ablation}.

\noindent \textbf{Geometric and Semantic Distillation.} 
Removing either the geometric ($4.16 \rightarrow 4.01$) or semantic ($4.16 \rightarrow 3.99$) distillation distinctly degrades performance. Lacking semantic foresight particularly harms long-horizon execution (the $5^{th}$ task success rate drops from $68.9\%$ to $64.1\%$), highlighting its critical role in maintaining consistent object identities to mitigate compounding errors.

\noindent \textbf{Self-Distillation.} To evaluate the necessity of the proposed mechanism, we compare our approach against a w/o Self-Distillation variant that uses VFM supervision extracted from ground-truth (GT) future frames instead of the model's own multi-step predictions. This variant degrades performance ($4.16 \rightarrow 3.82$). The resulting decline stems from the inherent trajectory misalignment between noisy one-step features and GT future representations, highlighting that self-distilling foresight from the model's own predictions is crucial for preserving a shared diffusion trajectory.

\noindent \textbf{Uni-Perceiver.} 
Removing the Uni-Perceiver causes a severe performance drop ($4.16 \rightarrow 3.72$). This demonstrates that adaptively condensing high-dimensional multi-modal features into a compact representation is essential for providing a holistic condition context to the action expert.

\noindent \textbf{Original Diffusion Features.} 
Relying exclusively on distilled foresight without the original one-step feature $F$ yields sub-optimal results ($4.16 \rightarrow 3.93$). The raw diffusion features are indispensable, as they supplement structural VFM representations by providing the action expert with residual global context.

\subsection{Alternative Vision Foundation Representations (Q3)}
\label{sec:ablation_vfm}

To demonstrate the effectiveness of our chosen combination of DINOv2~\cite{oquab2023dinov2} and DPAv3 as VFM targets, we evaluate a diverse set of alternative VFMs on the CALVIN~\cite{mees2022calvin} benchmark. We categorize them into semantic, geometric, motion-aware, and synergistic targets. The results are detailed in \cref{tab:ablation_vfm}.

\noindent \textbf{Semantic Targets (Dense vs. Global).} By extracting dense patch-level representations,
DINOv2~\cite{oquab2023dinov2} and DINOv3~\cite{simeoni2025dinov3} significantly outperform global semantic models CLIP~\cite{radford2021learning} and SigLIP~\cite{zhai2023sigmoid} (e.g., 4.01 for DINOv2 vs. 3.72 for CLIP in average length). This gap demonstrates that while global semantics provide scene-level context, precise manipulation inherently relies on dense semantic distinctiveness for the action expert to attend to task-relevant affordances. 

\noindent \textbf{Geometric Targets (Static vs. Dynamic).} 
DPAv3~\cite{lin2025depth} significantly outperforms VGGT~\cite{wang2025vggt} (3.99 vs. 3.74). We attribute this performance margin to DPAv3 being trained to adapt to dynamic video streams, enabling it to extract dynamic geometric structures. Conversely, VGGT focuses on static scene reconstruction, which makes it fundamentally unsuited to handle the dynamic object interactions inherent in robotic tasks.

\noindent \textbf{Motion-aware Targets.} 
We also explore using the representations extracted by video foundation models, VideoMAEv2~\cite{wang2023videomae} and V-JEPA2~\cite{assran2025v}, as motion-aware targets. However, they demonstrate sub-optimal results. We hypothesize that current video models still lag behind specialized image counterparts (e.g., DINOv2) in fine-grained feature fidelity.

\noindent \textbf{Synergistic Targets.} 
The best performance is achieved by fusing complementary modalities. Our choice (DINOv2 \& DPAv3) achieves the highest average sequence length of 4.16. Replacing DPAv3 with VGGT (DINOv2 \& VGGT) results in a performance drop to 4.04, confirming that static-scene geometric priors are insufficient for dynamic manipulation tasks.
Second, substituting DINOv2 with SigLIP (SigLIP \& DPAv3) leads to a decline to 4.06, demonstrating that dense and patch-level representations are more critical for low-level control than global-level language-aligned features.

\begin{table}[t]
\centering
\small
\caption{Quantitative comparison of different VFM representations as teacher targets.}
\label{tab:ablation_vfm}

\definecolor{mygray}{gray}{0.95}

\renewcommand{\arraystretch}{1.00}
\scalebox{0.96}{
\begin{tabular}{c|c|ccccc|c}
\toprule
\multirow{2}{*}{\textbf{Type}} & \multirow{2}{*}{\shortstack{\textbf{Representation} \\ \textbf{Source}}} & \multicolumn{5}{c|}{\textbf{$i^{th}$ Task Success Rate}} & \multirow{2}{*}{\textbf{Avg. Len.} $\uparrow$} \\
\cmidrule(lr){3-7}
 & & \textbf{1} & \textbf{2} & \textbf{3} & \textbf{4} & \textbf{5} & \\
\midrule

\multirow{4}{*}{\shortstack{Semantic \\ Targets}} 
 & CLIP~\cite{radford2021learning}      & 94.1 & 82.0 & 72.5 & 65.0 & 58.0 & 3.72 \\
 & SigLIP~\cite{zhai2023sigmoid}               & 94.5 & 84.8 & 74.7 & 65.5 & 57.0 & 3.77 \\
 & DINOv2~\cite{oquab2023dinov2}  & 94.1 & 87.1 & 79.3 & 73.5 & 66.5 & 4.01 \\
 & DINOv3~\cite{simeoni2025dinov3}               & 95.6 & 87.7 & 78.8 & 70.3 & 62.1 & 3.95 \\
\midrule

\multirow{2}{*}{\shortstack{Geometric \\ Targets}} 
 & DPAv3~\cite{lin2025depth} & 94.0 & 87.1 & 80.4 & 73.2 & 64.1 & 3.99 \\
 & VGGT~\cite{wang2025vggt}   & 93.5 & 84.3 & 74.8 & 65.6 & 55.8 & 3.74 \\ 

\midrule
\multirow{2}{*}{\shortstack{Motion-aware \\ Targets}} 
 & VideoMAEv2~\cite{wang2023videomae}  & 94.6 & 85.9 & 78.0 & 70.2 & 60.8 & 3.90 \\
 & V-JEPA2~\cite{assran2025v}  & 93.1 & 84.3 & 74.3 & 65.6 & 56.4 & 3.74 \\
\midrule

  & SigLIP \& DPAv3       & \textbf{96.2} & 88.3 & 80.7 & 74.2 & 66.1 & 4.06 \\
  & DINOv2 \& VGGT     & 95.7 & 88.8 & 80.9 & 73.9 & 64.9 & 4.04 \\
\rowcolor{mygray}
\cellcolor{white} \multirow{-3}{*}{\shortstack{Synergistic \\ Targets}} & \textbf{DINOv2 \& DPAv3 (our)} & {95.8} & {\textbf{90.7}} & {\textbf{83.7}} & {\textbf{77.0}} & {\textbf{68.9}} & {\textbf{4.16}} \\
\bottomrule
\end{tabular}
}
\end{table}

\subsection{Multi-task Experiments in the Real World (Q4)}
\label{sec:real_world_exp}
\begin{figure}[!t]
  \centering
  \includegraphics[width=0.99\linewidth]{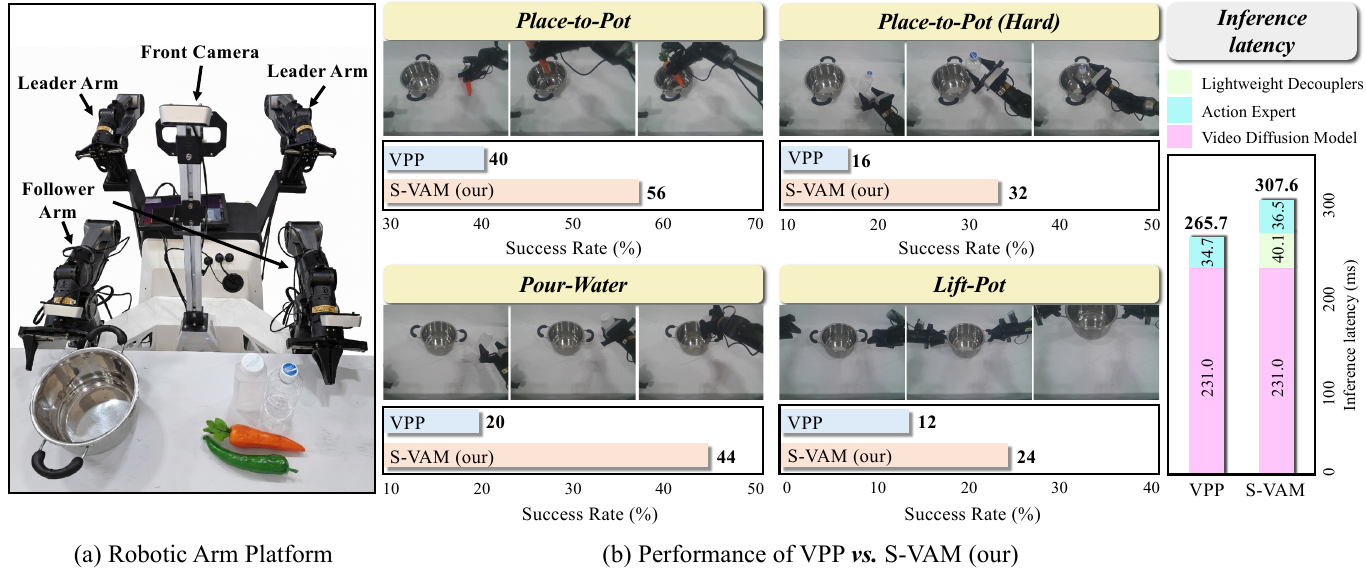}
 \caption{Multi-task real-world experiments. (a) We deploy our system on a dual-arm Cobot using only monocular front-camera observations. (b) Our S-VAM demonstrates a significant success rate improvement over VPP~\cite{huvideo} on all tasks without compromising real-time control capabilities.}
  \label{fig:real_world}
\end{figure}

\noindent \textbf{Task setups.}
As shown in \cref{fig:real_world}(a), we conduct our real-world experiments on a Cobot dual-arm robot manufactured by AgileX Robotics, which adopts the Mobile ALOHA system design~\cite{fu2024mobile}. Each arm possesses 7 degrees of freedom and is equipped with a parallel gripper. To maintain consistency with our simulated setup, all experiments are conducted using only the front camera to capture monocular RGB observations. We design four manipulation tasks to evaluate S-VAM's effectiveness: (1) \textit{Place-to-Pot}: picking an object and placing it into a pot; (2) \textit{Place-to-Pot (Hard)}: picking a transparent object and placing it into a pot; (3) \textit{Pour-Water}: pouring water from a transparent cup into a pot; and (4) \textit{Lift-Pot}: lifting the pot with both arms. We train a unified multi-task model on roughly 50 human demonstrations per task. During evaluation, we report the average success rate over 25 trials for each task.

\noindent \textbf{Results Analysis.}
As shown in \cref{fig:real_world}(b), our S-VAM outperforms VPP~\cite{huvideo} across all tasks. The performance gain is most evident in the \textit{Place-to-Pot (Hard)} task involving transparent objects. In this task, VPP achieves only 16\% success, as its noisy one-step features lack the coherent and stable foresight needed to resolve visual ambiguity. 
In contrast, our S-VAM benefits from the synergy of both decouplers: the geometric decoupler distills precise geometric priors to resolve the depth ambiguity of transparent surfaces, while the semantic decoupler extracts discriminative object features to maintain a consistent representation of the transparent entity. This distilled foresight enables the action expert to reliably anticipate the target's future states and plan precise action trajectories, boosting the success rate to 32\%. Furthermore, our S-VAM takes 307.6 ms per forward pass (video diffusion backbone: 231.0 ms; decouplers: 40.1 ms; action expert: 36.5 ms), introducing a modest 15.8\% overhead over VPP (265.7 ms). By predicting an action chunk of 8 per forward pass, our S-VAM achieves an effective control frequency of 25 Hz.
\section{Conclusion}

\label{sec:conclusion}

In this work, we present S-VAM, a novel framework that establishes a direct shortcut to resolve the dilemma between inference efficiency and foresight fidelity inherent to the video-action paradigm. Specifically, through a novel self-distillation strategy, we demonstrate that noisy one-step diffusion features can be successfully distilled into coherent geometric and semantic foresight, supervised by stable visual foundation representations extracted from multi-step generated videos. Crucially, this shortcut mechanism allows the policy to inherit high-fidelity planning while strictly maintaining the efficiency of single-step inference. Extensive experiments validate that S-VAM not only achieves state-of-the-art performance on simulated benchmarks but also unlocks efficient and precise manipulation capabilities in complex real-world scenarios.

\clearpage 
\bibliographystyle{splncs04}
\bibliography{main}
\end{document}